\documentclass[a4paper,fleqn]{cas-dc}
\pdfoutput=1

\usepackage{placeins}
\usepackage{float}
\usepackage{times}
\usepackage{epsfig}
\usepackage{graphicx}
\usepackage{amsmath}
\usepackage{amssymb}
\usepackage{tikz}
\usepackage{multirow}
\usepackage{algpseudocode}
\usepackage{algorithm}

\usepackage[numbers]{natbib}

\def\tsc#1{\csdef{#1}{\textsc{\lowercase{#1}}\xspace}}
\tsc{WGM}
\tsc{QE}
\tsc{EP}
\tsc{PMS}
\tsc{BEC}
\tsc{DE}

\newcommand\restr[2]{{
  \left.\kern-\nulldelimiterspace 
  #1 
  \right|_{#2} 
  }}

\newtheorem{theorem}{Theorem}

\newtheorem{lemma}{Lemma}
\newtheorem{remark}{Remark}
\newtheorem{example}{Example}
\usepackage{caption}
\usepackage{subcaption}

\DeclareMathOperator{\argmax}{arg\,max}
\newcommand{\R}{\mathbb{R}}
\newcommand{\N}{\mathcal{N}}
\newcommand{\Pp}{\mathcal{P}}
\newcommand{\Ll}{\mathcal{L}}
\newcommand{\M}{\mathcal{M}}

\DeclareMathOperator{\Del}{Del}
\DeclareMathOperator{\softmax}{softmax}

\newcommand{\proof}{\noindent{\bf Proof. }}

\begin{document}

\let\WriteBookmarks\relax
\def\floatpagepagefraction{1}
\def\textpagefraction{.001}

\shorttitle{Trainable and Explainable  SMNNs}
\shortauthors{Paluzo-Hidalgo E., Gonzalez-Diaz R., Guti\'errez-Naranjo M. A.}

\title[mode = title]{Trainable and Explainable Simplicial Map Neural Networks}                      



\author[1]{Eduardo Paluzo-Hidalgo}[orcid=0000-0002-4280-5945]
\cormark[1]
\ead{epaluzo@us.es}
\address[1]{Department of Quantitative Methods, Universidad Loyola Andalucía, Campus Sevilla, Dos Hermanas, Seville, Spain}
\author[2]{Rocio Gonzalez-Diaz}[orcid=0000-0001-9937-0033]
\ead{rogodi@us.es}
\ead[url]{https://personal.us.es/rogodi}
\address[2]{Department of Applied Mathematics I, School of engineering, University of Seville, Seville, Spain}
\author[3]{Miguel A. Guti\'errez-Naranjo}[orcid=0000-0002-3624-6139]
\ead{magutier@us.es}
\ead[URL]{http://www.cs.us.es/~naranjo/}
\address[3]{Department of Computer Science and Artificial Intelligence, School of Engineering, University of Seville, Seville, Spain}
\cortext[cor1]{Corresponding author}

\begin{abstract}
Simplicial map neural networks (SMNNs) are topology-based neural networks with interesting properties such as universal approximation ability 
and robustness to adversarial examples under appropriate conditions.
However, SMNNs present some bottlenecks for their possible application in high-dimensional datasets.
First, SMNNs have precomputed fixed weight and no SMNN training process has been defined so far, so they lack generalization ability. Second, SMNNs require the construction of a convex polytope surrounding the input dataset.
In this paper, we overcome these issues by proposing an SMNN training procedure based on a support subset of the given dataset and replacing the construction of the convex polytope by a method based on projections to a hypersphere.
In addition, the explainability capacity of SMNNs and effective implementation are also newly introduced in this paper.
\end{abstract}

\begin{keywords}
Training neural network \sep Simplicial maps \sep  
\sep Explainable artificial intelligence 
\end{keywords}

\maketitle

\section{Introduction}

In recent years, Artificial Intelligence (AI) methods in general and Machine Learning methods in particular have reached success in real-life problems that were unexpected only a few years ago. Many different areas have contributed to this development. Among them, we can cite the research on new theoretical algorithms, the increasing computational power of the latest generation hardware, and the rapid access to a huge amount of data. Such a combination of factors leads to the development of increasingly complex self-regulated AI methods.

Many AI models currently used are based on backpropagation algorithms, which train and regulate themselves to achieve a goal, such as classification, recommendation, or prediction. These self-regulating models achieve some kind of knowledge as they successfully evaluate test data independent of the data used to train them. Nonetheless, such knowledge is usually expressed in a non-human-readable way.
 
To fill the gap between the recent development of AI models and their social use, many researchers have focused on the development of Explainable Artificial Intelligence (XAI), which consists of a set of techniques to provide clear, understandable, transparent, intelligible, trustworthy, and interpretable explanations of the decisions, predictions, and reasoning processes made by the AI models, rather than just presenting their output, especially in domains where AI decisions can have significant consequences on human life.
 A global taxonomy of interpretable AI with the aim of unifying terminology to achieve clarity and efficiency in the definition of regulations for the development of ethical and reliable AI can be found in \cite{Gra22}. 
Moreover, a nice introduction and general vision can be found in \cite{Mol22}. Another clarifying paper with definitions, concepts, and applications of XAI is \cite{MSQ19}.

The so-called Simplicial Map Neural Networks (SMNNs) were introduced in \cite{DBLP:journals/nn/Paluzo-HidalgoG20} as a constructive approach to the problem of approximating a continuous function on a compact set in a triangulated space. Since the original aim of the definition of SMNNs was focused on building a {\it constructive} approach, the computation of their weights was not based on an optimization process, as usual in neural networks, but on a deterministic calculus. The architecture of SMNNs and the computation of the set of weights are based on a combinatorial topology tool called simplicial maps. 
Moreover, SMNNs can be used for classification and can be constructed to be robust to adversarial examples \cite{smnn}. 
Besides,  their architecture can be reduced while maintaining accuracy \cite{smnn-optimo}, being invariant to transformation if the transformation preserves the barycentric coordinates (scale, rotation, symmetries, etc.).
As defined in \cite{DBLP:journals/nn/Paluzo-HidalgoG20,smnn-optimo,smnn},  SMNNs are built as a two-hidden-layer feed-forward network where the set of weights is precomputed based on the calculation of a triangulated convex polytope surrounding the input data.
As other approximations to continuous functions with arbitrary precision (see, for example, \cite{uno}), SMNNs have fixed weights, which means
that the weights depend only on the triangulation made with the points of the dataset as the  {\it support} set and no training process is applied.

Summing up, some of the limitations of the SMNNs until now are that they are costly to calculate since the number of neurons is proportional to the number of simplices of the triangulation supported on the input dataset, and they suffer from overfitting and therefore not generalize well. These aspects make SMNNs not used in practice so far, although the idea of relating simplicial maps to neural networks is disruptive and provides a new bridge that can enrich both areas.

In this paper, we propose a method to make SMNNs efficient by reducing
their size (in terms of the number of neurons that depends on the vertices of the triangulation) and that successfully makes SMNNs trainable and with generalization ability.
Besides, we also present a study of the selection of the vertices from which we obtain the triangulation.
Although SMNNs consider the vertices of a simplex as part of the necessary information for the classification task, the approach presented in this paper is far from the classic Machine Learning instance-based methods. Such methods rely on a deterministic computation based on distances, but, in the approach presented in this paper, the computation of the weights is the result of an optimization method in a probability distribution space.
Finally, from an XAI point of view, we will see in this paper that SMNNs are explainable models since all decision steps to compute the output of SMNNs are understandable and transparent, and therefore trustworthy.

The paper is organized as follows. First, some concepts of computational topology and the definition of SMNNs are recalled in Section~\ref{sec:back}. Next, in Section~\ref{sec:unknown} we develop several technical details needed for the SMNN training process, which will be introduced in Section~\ref{sec:train}. 
Section~\ref{sec:Explain} is devoted to the explainability of the model. 
Section \ref{sec:disc} is devoted to discussion and limitations.
Finally, the paper ends with some experiments and conclusions.


\section{Background}\label{sec:back}
In this section, we assume that the reader is familiar with the basic concepts of computational topology. For a comprehensive presentation, we refer to \cite{books/daglib/0025666}.

\subsection{Simplicial complexes}\label{subsec:simplicial_complex}

Consider a finite set of points $V=\{v^1,\dots,v^{\beta}\}\subset \mathbb{R}^n$
whose elements will be called vertices. 
A subset 
$$\sigma=\langle v^{i_0},v^{i_1},\dots,v^{i_d}\rangle$$ of $V$ with $d+1$ vertices (in general position) is called a $d$-simplex.  
The convex hull of the vertices of $\sigma$ will be denoted by $|\sigma|$ and corresponds to the set: 
$$\mbox{$\Big\{x\in\mathbb{R}^n:\, x=\sum_{\scriptscriptstyle j\in [\![ 0,d]\!]}b_j(x)v^{i_j}\Big\}$}$$
where $[\![ a,b]\!]=\{a,a+1,\dots,b\}$ for $a<b\in\mathbb{Z}$,
and
  $$b(x)=(b_0(x),b_1(x),\dots,b_d(x))$$ are  called the barycentric coordinates of $x$ with respect to $\sigma$, and  satisfy that:
$$\mbox{$\sum_{\scriptscriptstyle j\in [\![ 0,d]\!]} b_j(x)=1\;\mbox{ and }\; b_j(x)\geq 0\; \forall j\in [\![ 0,d]\!]$}\,.$$
The barycentric coordinates of $x$ can be interpreted as masses placed at the vertices of  $\sigma$ so $x$ is the centre of mass. All these masses are positive if and only if $x$ is inside $\sigma$. For example, let us consider the 1-simplex
$\epsilon=\langle v^{i_0},v^{i_1}\rangle$ which is composed of two vertices of $V$.
 Then $|\epsilon|$ is the set of points in $\R^n$ corresponding to the edge with endpoints $v^{i_0}$ and $v^{i_1}$, and if, for example, $b(x)=(0.5,0.5)$ then $x$ is the midpoint of $|\epsilon|$.

\begin{figure}
\begin{center}%
\includegraphics[width=\linewidth]{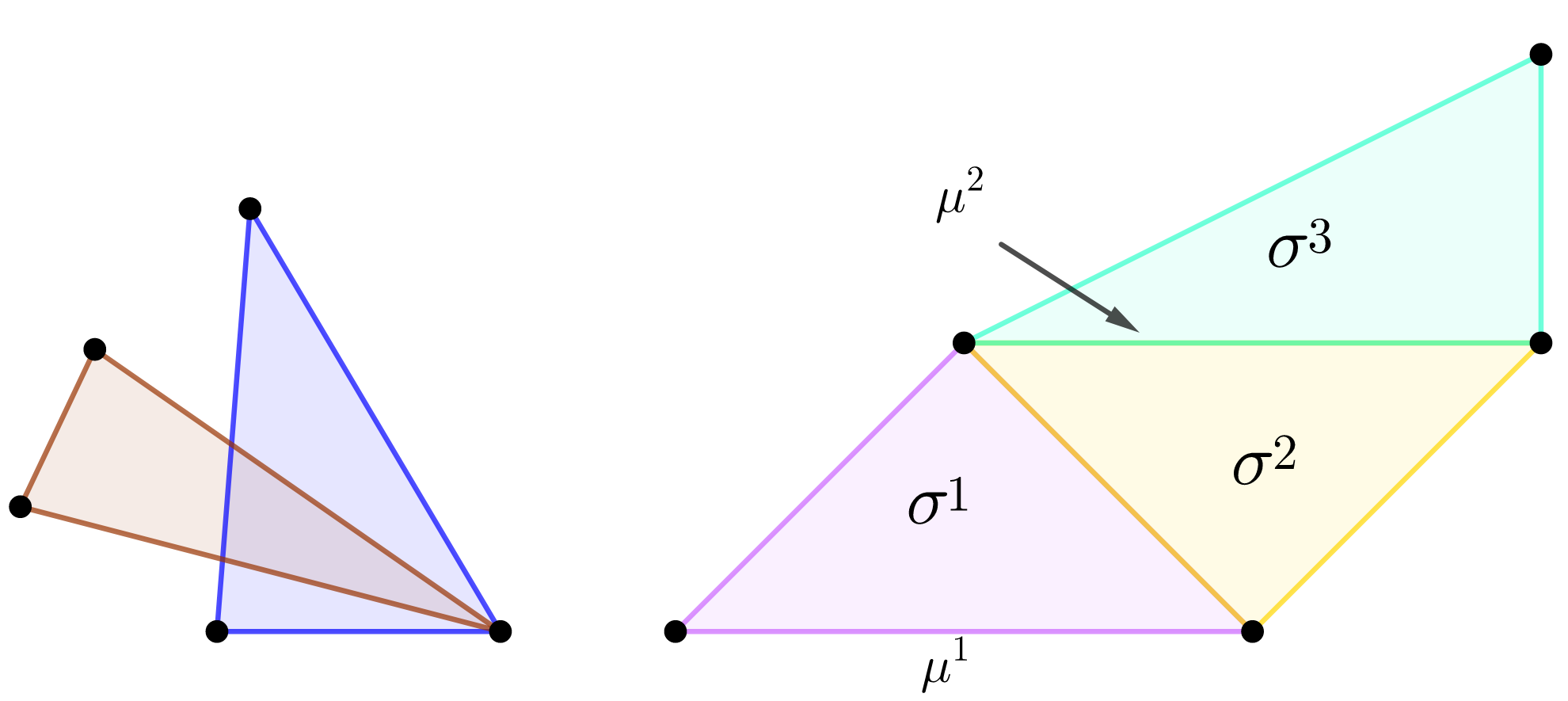}
\end{center}
\caption{On the left, two triangles that do not intersect in a common face (an edge or a vertex). 
On the right, the geometric representation $|K|$ of a pure 2-simplicial complex $K$ composed of three maximal $2$-simplices (the triangles $\sigma^1$, $\sigma^2$ and $\sigma^3$). The edge $\mu^2$ is a common face of $\sigma^2$ and $\sigma^3$. The edge $\mu^1$ is a face of $\sigma^1$. }
\label{fig:face}
\end{figure}

A simplicial complex $K$ with vertex set $V$ consists of a finite collection of simplices satisfying that if $\sigma \in K$ then either $\sigma=\langle v\rangle$ for some $v\in V$ or any face (that is, a nonempty subset) of $\sigma$  is a simplex of $K$. Furthermore, if $\sigma,\mu\in K$ then $|\sigma|\cap |\mu|=\emptyset$ or $|\sigma|\cap |\mu|=|\gamma|$ for some $\gamma\in K$. 
The set $\bigcup_{\sigma\in K}|\sigma|$ will be denoted by $|K|$.
A maximal simplex of $K$ is a simplex that is not the face of any other simplex of $K$. 
If the maximal simplices of $K$ are all $d$-simplices then $K$ is called a pure $d$-simplicial complex.
These concepts are illustrated in Figure~\ref{fig:face}.

The barycentric coordinates of $x$ with respect to the simplicial complex $|K|$ are defined as the barycentric coordinates of $x$ with respect to $\sigma\in K$ such that $x\in|\sigma|$.
Let us observe that the barycentric coordinates of $x\in |K|$ are unique. 

An example of simplicial complexes is the Delaunay triangulation $\Del(V)$ defined from the Voronoi diagram of a given finite set of vertices $V$. The following result
extracted from \cite[page 48]{boissonnat_chazal_yvinec_2018}  is just an alternative definition of Delaunay triangulations.
\\
\\
\noindent{\bf 
The empty ball property
\cite{boissonnat_chazal_yvinec_2018}:}
Any subset $\sigma$ of $V$ is a simplex of  $\,\Del(V)$  if and only if $|\sigma|$ has a circumscribing open ball empty of points of $V$.

\subsection{Simplicial maps}\label{subsec:simp-map}

Let $K$ be a pure $n$-simplicial complex and $L$ a pure $k$-simplicial complex with vertex sets $V$ and $W$, respectively.
  The map $\varphi^{\scriptscriptstyle (0)}:V\to W$ is called a {\it vertex map} if it satisfies that the set obtained from $\big\{ \varphi^{\scriptscriptstyle (0)}(v^{i_0}),\dots,\varphi^{\scriptscriptstyle (0)}(v^{i_d})\big\}$ after removing duplicated vertices is a simplex in $L$ whenever $\langle v^{i_0},\dots,v^{i_d}\rangle$ is a simplex in $K$.
The vertex map $\varphi^{\scriptscriptstyle (0)}$ always induces a continuous function, called a {\it simplicial map} $\varphi:|K|\to|L|$, which is defined as follows. Let $b(x)=(b_0(x),\dots, b_n(x))$ be the barycentric coordinates of $x\in |K|$ with respect to $\sigma= \langle v^{i_0},\dots,v^{i_n}\rangle\in K$.
Then 
$$\mbox{$\varphi(x)=\sum_{\scriptscriptstyle j\in [\![ 0,n]\!]} b_j(x)\varphi^{\scriptscriptstyle (0)}(v^{i_j})$}.$$
Let us observe that  
$\varphi(x)=\varphi^{\scriptscriptstyle (0)}(x)$ if $x\in V$.

A special kind of simplicial map used to solve classification tasks will be introduced in the next subsection.

\subsection{Simplicial maps for classification tasks}\label{subsubsec:maps}

Next, we will show how a simplicial map can be used to solve a classification problem (see \cite{smnn} for details). From now on, we will assume that the input dataset is a finite set of points $V$ in $\R^n$ together with a set of $k$ labels $\Lambda$ such that each $v\in V$ is tagged with a label $\lambda$
 taken from $\Lambda$. 

Firstly, the intuition is that the space surrounding the dataset is labelled as {\it unknown}. 
For this, we add a new label to $\Lambda$, called {\it unknown} label, and 
a one-hot encoding representation $W^{k+1}\subset\R^{k+1}$ of these $k+1$ labels being:
 $$W^{k+1}=  \big\{\ell^j=(0,\stackrel{j}\dots,0,1,0,\stackrel{k-j}{\dots},0):\;  j\in [\![1,k]\!]\big\}\,,  $$
 where the one-hot vector $\ell^j$ encodes the $j$-th label of $\Lambda$ for $j\in [\![1,k]\!]$ and  where $\ell^0$ encodes the {\it unknown} label. 

We now consider a convex polytope $\Pp$ with a vertex set $P$ surrounding the set $V$. The polytope $\Pp$ always exists since $V$ is finite. 
Next, we define a map $\varphi^{\scriptscriptstyle (0)}:V\cup P\to W^{k+1}$ mapping each vertex $v\in V$ tagged with label $\lambda$ to the one-hot vector in $W^{k+1}$ that encodes the label $\lambda$.
The vertices of $P$ are sent to the vertex $\ell^0$.
Observe that $\varphi^{\scriptscriptstyle (0)}$ is a vertex map.  

  Let $L$ denote the simplicial complex with vertex set $W^{k+1}$ consisting of only one maximal $k$-simplex   and let  $\Del(V\cup P)$ denote the Delauney triangulation computed for the set of points $V\cup P$.
Note that $|\Del(V\cup P)|=\Pp$.
The simplicial map $\varphi:\Pp\to |L|$  is induced by the vertex map $\varphi^{\scriptscriptstyle (0)}$ as explained in Subsection~\ref{subsec:simp-map}. 

\begin{remark}
The space $|L|$ can be interpreted as the discrete probability distribution space $\Omega^{k+1}$ with $k+1$ variables.
 \end{remark}

\begin{figure}
\begin{center}
\includegraphics[width=0.45\textwidth]{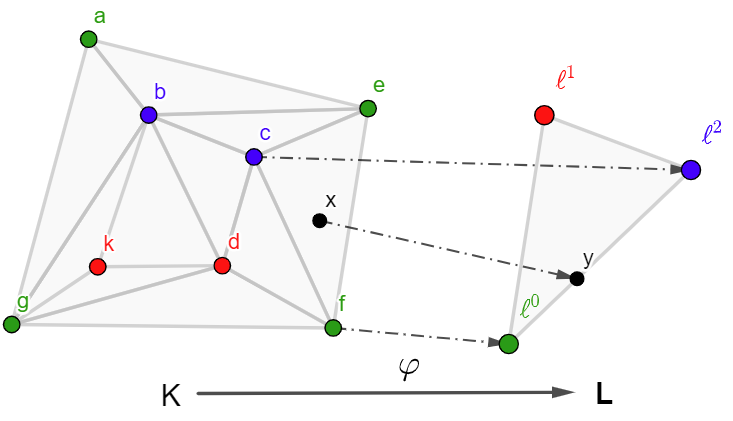}   
\end{center}
\caption{Illustration of a simplicial map for a classification task.}
\label{Fig:simplicial_map}
\end{figure}

As an example, in Figure~\ref{Fig:simplicial_map}, on the left, we can see a dataset with four points $V=\{b,c,k,d\}$, labelled red and blue. 
The green points $P=\{a,e,g,f\}$ are the vertices of a convex polytope $\Pp$ containing $V$ and are sent by $\varphi^{\scriptscriptstyle (0)}$ to the green vertex $\ell^0$ on the right. 
The simplicial complex $K=\Del(V\cup P)$ is drawn on the left and consists of ten maximal 2-simplices.  
On the right, the simplicial complex $L$ consists of one maximal 2-simplex. The dotted arrows illustrate some examples of $\varphi:\Pp\to |L|$.

\subsection{Simplicial map neural networks}\label{sec:smnn}

Artificial neural networks can be seen as parametrized real-valued mappings between multidimensional spaces. Such mappings are the composition of several maps (usually many of them) that can be structured in layers. 
In \cite{smnn}, the simplicial map $\varphi$ defined above was represented as a two-hidden-layer feed-forward neural network $\N_{\varphi}$. 
This kind of artificial neural network is called {\it simplicial map neural network} (SMNN). 

In the original definition \cite{smnn}, the first hidden layer of an SMNN computes the barycentric coordinates of $\Del(V\cup P)$.
However, we will see here that if we precompute the barycentric coordinates, we can simplify the architecture of $\N_{\varphi}$ as follows. 

As before, consider an input dataset consisting of a finite set $V\subset \R^n$ endowed with a set of $k$ labels and a convex polytope $\Pp$ with a set of vertices $P$ surrounding $V$. Let $\Del(V\cup P)$ be the Delaunay triangulation with vertex set  
$$V\cup P=\{\omega^1,\dots,\omega^\alpha\}\subseteq \R^n\,.$$ 
Then, $\varphi^{\scriptscriptstyle (0)}:V\cup P\to W^{k+1}$ is a vertex map. Let us assume that, given $x\in\Pp$, we precompute the barycentric coordinates $b(x)=(b_0(x),\dots,b_n(x))\in\R^{n+1}$ of $x$  with respect to the $n$-simplex $\sigma=\langle \omega^{i_0},\dots,\omega^{i_n}\rangle\in \Del(V\cup P)$ such that $x\in|\sigma|$, and that we also precompute the vector 
$$\xi(x)=(\xi_1(x),\dots,\xi_\alpha(x))\in\R^\alpha$$ 
satisfying that, 
 for $t\in [\![1,\alpha]\!]$, $\xi_t(x)=b_j(x)$ if
 $i_j=t$ for some $j\in [\![0,n]\!]$.
 Let us remark that $\xi(x)$ should be a column vector, but we will use row notation, for simplicity.

The SMNN $\N_{\varphi}$ induced by $\varphi$ that predicts the $h$-label of $x$, for $h\in [\![0,k]\!]$, has the following architecture:
\begin{itemize}
    \item The number of neurons in the input layer is $\alpha$.
    \item The number of neurons in the output layer is $k+1$.
    \item The set of weights is represented as a $(k+1)\times \alpha$ matrix $\M$ such that
    the $j$-th column of $\M$ is 
 $\varphi^{\scriptscriptstyle  (0)}(\omega^t)$ for $t\in[\![1,\alpha]\!]$.
    \end{itemize}
Then,  
$$ \N_{\varphi}(x)=\M\cdot \xi(x)\ .$$

Observe that as defined so far, the SMNN weights are precomputed.
 Furthermore, the computation of the barycentric coordinates of the points around $V$ implies the calculation of the convex polytope $\Pp$ surrounding $V$. 
Finally, the computation of the Delaunay triangulation $\Del(V\cup P)$ is costly if $V\cup P$ has many points since its time complexity is
$O(n\log{n}+n^{\lceil \frac{d}{2}\rceil })$ (see \cite[Chapter 4]{boissonnat_chazal_yvinec_2018}).

In the next sections, we will propose some techniques to overcome the SMNN construction drawbacks while maintaining its advantages. 
We will see that one way to overcome the computation of the convex polytope $\Pp$ is to consider a hypersphere $S^n$ instead. 
We will also see how to avoid the use of the artificially created {\it unknown} label. 
Furthermore, to reduce the cost of Delaunay computation and add trainability to $\N_{\varphi}$ to avoid overfitting, a subset $U\subset V$ will be considered. 
The set $V$ will be used to train and test a map $\varphi_{\scriptscriptstyle U}^{\scriptscriptstyle (0)}: U\to\R^{k}$. 
Such a map will induce a continuous function $\varphi_{\scriptscriptstyle U}: B^n\to |L|$ which approximates $\varphi$.

\section{The {\it unknown} boundary and the function $\varphi_{\scriptscriptstyle U}$}\label{sec:unknown}

 In this section, we will see how to compute a function $\varphi_{\scriptscriptstyle U}$ that approximates the simplicial map $\varphi$ and avoids the computation of the convex polytope $\Pp$ and the artificial consideration of the {\it unknown} label, reducing, at the same time, the computation of the Delaunay triangulation used to construct SMNNs. 
 The general description of the methodology is described in Algorithm~\ref{algo:SMNN}.

First,  let us compute a hypersphere surrounding $V$. 
One of the simplest ways to do that is to translate $V$ so that its center of mass is placed at the origin $o\in\R^n$. 
Then, the hypersphere
\[
S^n=
\{w\in \R^n:\; ||w||= R\}
\]
such that $R>\max\{||v||:\, v\in V\}$ satisfies that $S^n$ surrounds $V$.
Second, let us assume that we have selected a subset $U=\{u^1,\dots,u^m\}\subseteq V$ (we will compare different strategies to select $U$ in Section \ref{sec:experiments}) and that we have computed $\Del(U)$. 
Then, we have that
\[
V\subset  B^n=\{x\in \R^n:\; ||x||\leq R\}\;\mbox{ and }\;o\in|\Del(U)|\,.
\]
Let us consider the boundary of $\Del(U)$, denoted as $\delta\Del(U)$, which consists of the set of $(n-1)$-simplices that are faces of exactly one maximal simplex of $\Del(U)$.  

Now,  let us define $\xi_U(x)\in\R^m$ for any $x\in B^n$ as follows. 
Given $x\in B^n$, to find the $n$-simplex $\sigma=\langle \omega^0,\dots,\omega^n\rangle$ such that $x\in|\sigma|$, we have to consider two cases:   $x\in|\Del(U)|$ and $x\not\in|\Del(U)|$.

 If $x\in|\Del(U)|$ then $\sigma$ is the $n$-simplex in $\Del(U)$ such that $x\in|\sigma|$.
 If  $x\not\in|\Del(U)|$ then $\sigma$ is a new $n$-simplex  defined by the vertices of an $(n-1)$-simplex of $\delta\Del(U)$ and a new vertex consisting of the projection of $x$ to $S^n$.
   Specifically, if $x\not\in|\Del(U)|$ then $\sigma$ is computed in the following way:
\begin{enumerate}
\item Consider the set 
$$\Gamma=\big\{\mu\in \delta\Del(U):
(N \cdot u^{i_0} + c)(N \cdot x + c)<0\big\}$$
    where
 $N$ is the  vector normal to the hyperplane
     containing 
     $\mu=\langle u^{i_1},\dots,u^{i_n}\rangle$,
     $c=N\cdot u^{i_1}$,
       and   
    $u^{i_0}\in U$ such that $\langle u^{i_0},u^{i_1},\dots,u^{i_n}\rangle\in\Del(U)$.

      \item Compute the point $w^x=R\frac{x}{||x||}\in S^n$. 
\item Find $\sigma=\langle w^x,u^{i_1},\dots,u^{i_n}\rangle$ such that 
\[\mbox{$\mu=\langle u^{i_1},\dots,u^{i_n}\rangle\in \Gamma\;$ and $\;x\in|\sigma|$.}\]
Observe that, by construction, $\mu$ 
always exists since $|\Del(U)|$ is a convex polytope.
\end{enumerate}

Now, let $b(x)=(b_0(x),\dots,b_n(x))\in\R^{n+1}$ be the barycentric coordinates of $x$ with respect to $\sigma$.
Then,
$\xi_U(x)=(\xi_1(x),\dots,\xi_m(x))$ is the point in $\R^m$ satisfying that, 
 for $t\in [\![1,m]\!]$, 
 $$\xi_t(x)=\left\{ 
 \begin{array}{cl}
 b_j(x) &\mbox{ if 
 $u^t=\omega^j$ for some $j\in [\![0,n]\!]$, } 
 \\0 &\mbox{ otherwise.}\end{array}\right.$$

\begin{figure}
\begin{center}
\includegraphics[width=0.7\linewidth]{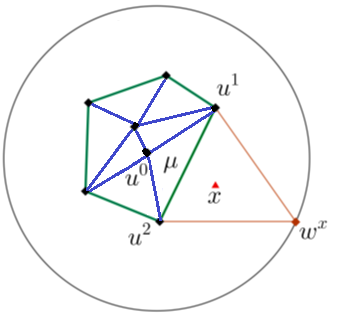}
\end{center}
\caption{An example of the point $w^x$ computed from $x$ and the $(n-1)$-simplex $\mu=\langle u^{1},u^{2}\rangle\in \Gamma$ such that
$x\in|\sigma|$ for $\sigma=\langle w^x,u^{1},u^{2}\rangle$.}
\label{fig:wx}
\end{figure}

Observe that $\xi_U(x)$ always exists and is unique. An example of points $x$ and $w^x$ and simplex $\mu$ is shown in Figure~\ref{fig:wx} and Example~\ref{example}.

Let us observe that, thanks to the new definition of $\xi_U(x)$ for $x\in B^n$, if we have a map 
 $\varphi_{\scriptscriptstyle U}^{\scriptscriptstyle (0)}:U\to \R^k$
 then it induces a continuous function $\varphi_{\scriptscriptstyle U}: B^n\to |L|$ defined for any $x\in B^n$ as:
$$\begin{array}{cl}
\varphi_{\scriptscriptstyle U}(x)&=\softmax\big(\sum_{\scriptscriptstyle t\in [\![ 1,m]\!]} \xi_t(x)\varphi_{\scriptscriptstyle U}^{\scriptscriptstyle (0)}(u^t)\big)\\
&=\softmax(\M_U\cdot\xi_U(x))
\end{array}$$
where for $z=(z_1,\dots,z_k)\in\R^{k}$,
\[\mbox{$\softmax(z)= \left(\frac{e^{z_1}}{\sum_{\scriptscriptstyle h\in [\![1,k]\!]} e^{z_h}},\dots, \frac{e^{z_k}}{\sum_{\scriptscriptstyle h\in [\![1,k]\!]} e^{z_h}}\right)$}.\]
Let us observe that, to obtain a categorical distribution from  $\varphi_{\scriptscriptstyle U}(x)\in \R ^k$, we could just divide each of its coordinates by the total sum. However, $\softmax$ is introduced here to obtain a simplified formula for the gradient descent algorithm as shown in Theorem~\ref{th:1}.

\begin{figure}
    \centering
\includegraphics[width=0.4\textwidth]{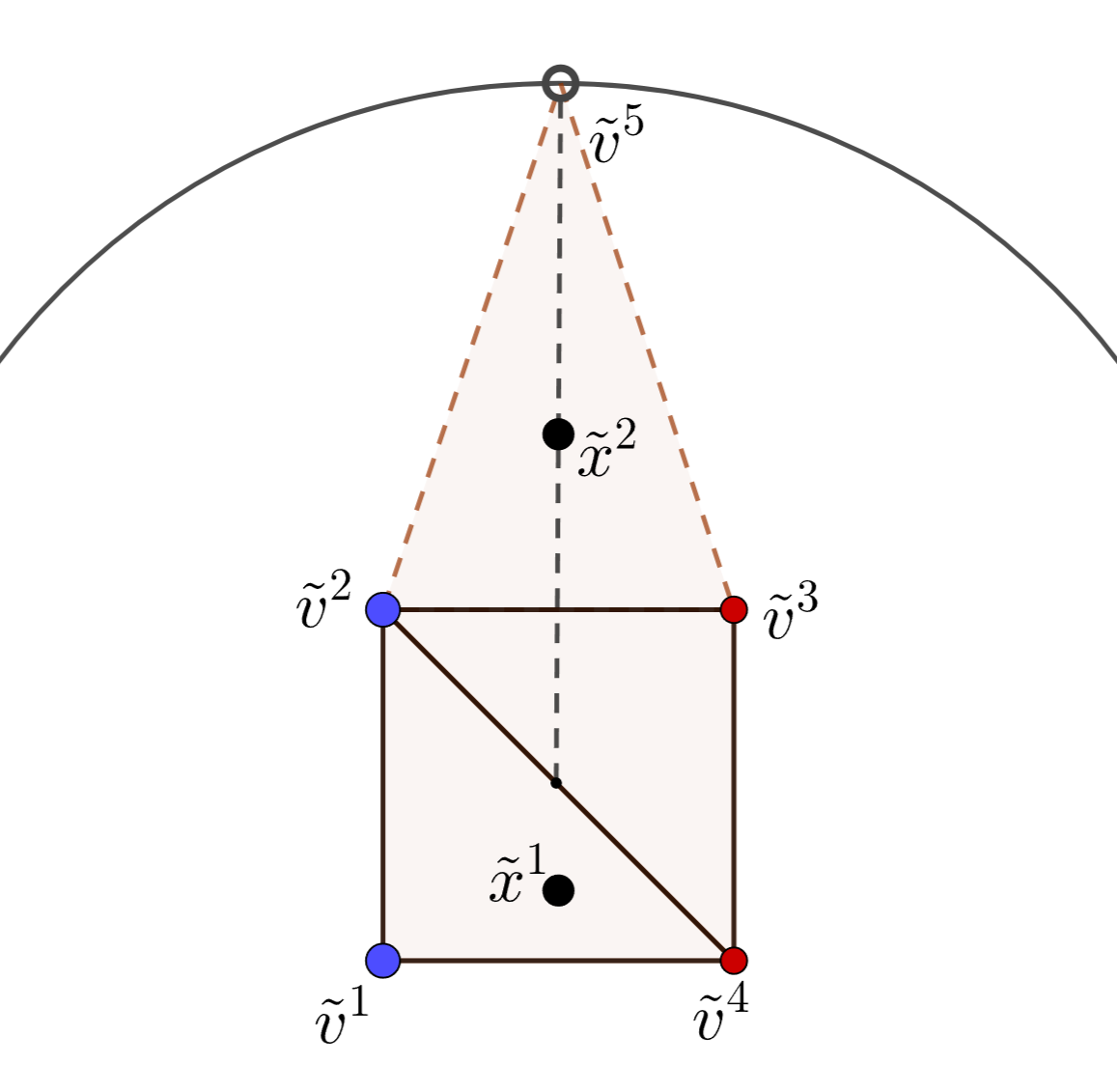} 
    \caption{
    The relative positions of the vertices $\tilde{v}^i$ for $i\in [\![1,5]\!]$     and the points $\tilde{x}^1$ and $\tilde{x}^2$ of Example~\ref{example}.}
     \label{fig:example_hand}
\end{figure}

\begin{example}\label{example}
Let us consider 
$$\mbox{$V=\big\{v^1=\big(\frac{1}{2},\frac{1}{2}\big),\,v^2=\big(\frac{1}{2},1\big),\,v^3=\big(1,\frac{1}{2}\big),\,v^4=\big(1,1\big)\big\}$}$$ 
with label $\lambda^1=0$ for $v^i$ with $i=1,2$ and $\lambda^2=1$ 
for $v^i$ with $i=3,4$.
Firstly, we translate $V$ so that the center of mass of $V$ is the origin $o\in\R^2$.
The translated dataset $\tilde{V}$ is 
$$\mbox{$\big\{\tilde{v}^1=\big(\frac{-1}{4},\frac{-1}{4}\big),\tilde{v}^2=\big(\frac{-1}{4},\frac{1}{4}\big),\tilde{v}^3=\big(\frac{1}{4},\frac{-1}{4}\big),\tilde{v}^4=\big(\frac{1}{4},\frac{1}{4}\big)\big\}$}.$$ 
Let us consider $x^1=\big(\frac{3}{4},\frac{3}{5}\big)$ and $x^2=\big(\frac{3}{4},\frac{5}{4}\big).$
Hence, the translated input data is $\tilde{x}^1=\big(0,\frac{-3}{20}
\big)$ and $\tilde{x}^2=\big(0,\frac{1}{2}\big)$.
\\
To simplify the explanation of the method, consider $U=\tilde{V}$, that is, $u^i=\tilde{v}^i$ for $i\in [\![1,4]\!]$.
Then, the maximal simplices of $\Del(U)$ are $\tilde{\sigma}^1=\langle \tilde{v}^1,\tilde{v}^2,\tilde{v}^3\rangle$ and
$\tilde{\sigma}^1=\langle\tilde{v}^2,\tilde{v}^3,\tilde{v}^4\rangle$ 
\begin{itemize}
\item The matrix $\mathcal{M}_U$ is:
     $\begin{pmatrix}
  1 & 1 & 0 &0\\
      0 & 0 &1 &1\\
     \end{pmatrix}$
    \item Since the barycentric coordinates of $\tilde{x}^1$ with respect to $|\tilde{\sigma}^1|$ are $(0.5,0.3,0.2)$ then $\tilde{x}^1$ is in $|\tilde{\sigma}^1|\subset |\Del(U)|$ and $\xi_U(\tilde{x}^1)=(0.3,0.2,0,0.5)$.
    Then
    $$\varphi_U
    (\tilde{x}^1)
    =\softmax(\M_U\cdot\xi_U(\tilde{x}^1))
    =(0.5,0.5).$$
    \item On the other hand, the point $\tilde{x}^2$ is outside
    $|\Del(U)|$.
   Assuming that, for example, we have fixed $R=1$, we add a new simplex $\sigma^3=\{w^x,\tilde{v}^1,\tilde{v}^2\}$ where $w^x=\tilde{v}^5=(0,1)$ which is the projection of $\tilde{x}^2$ to the hypersphere of radius $R$ centered in the origin. See Figure~\ref{fig:example_hand}.
    Then, the barycentric coordinates of $\tilde{x}^2$ with respect to $\sigma^3$ is $(0.33,0.33,0.33)$ and then $\xi_U(\tilde{x}^2)=(0,0.33,0.33,0)$, concluding that
        $$\varphi_U(\tilde{x}^2)=\softmax(\M_U\cdot\xi_U(\tilde{x}^2))=(0.5,0.5).$$ 
\end{itemize}
\end{example}

The pseudocode for computing $\varphi_U(x)$ is provided in Algorithm~\ref{algo:SMNN}.

\begin{algorithm}
\caption{Pseudocode to compute $\varphi_U(x)$ for $x\in B^n$ and a subset $U$ of an input dataset $V$ surrounded by a hypersphere of radius $R$}
\label{algo:SMNN}
\begin{algorithmic}
\State {\bf Input:} 
$U\subset V\subset \mathbb{R}^n$ labelled using a set of labels $\Lambda=\{\lambda^1,\dots,\lambda^k\}$, a radius $R$, and a point $x\in B^n$.
\State {\bf Output:} The value of $\varphi_U(x)$
\State {\bf compute} $\Del(U)$
\State  $W^{k}:=
\big\{\ell^j:=(0,\stackrel{j-1}\dots,0,1,0,\stackrel{k-j}{\dots},0)$ for $j\in[\![1,k]\!]\big\}$
\State {\bf init} an empty matrix $\M_U$
\State {\bf init} an empty vector $xi(x)$
 \For{$u\in U$}
 \If{$\lambda^j$ is the label of $u$} 
\State {\bf add} $\ell^j$ as a column of $\mathcal{M}_U$ 
\EndIf
\EndFor
 \For{$\sigma$ maximal simplex of $\Del(U)$}
 \State {\bf compute} $b(x)$ 
 \If{$b_j\geq 0$ for all $j\in [\![0,n]\!]$}
 \State {\bf stop:} {\bf compute} $\xi_U(x)$
 \EndIf
 \EndFor
\If{$\xi_U(x)$ is empty}
\State {\bf compute} $\Gamma$, $w^x$, $\mu$ and $\sigma$ 
\State {\bf compute} $b(x)$ with respect to $\sigma$ and $\xi_U(x)$
\EndIf
\State $\varphi_U(x)
 :=\softmax (\M_U\cdot \xi_U(x))$
\end{algorithmic}
\end{algorithm}

The following property holds.

\begin{lemma}[Continuity]
Let $x\in B^n$. Then,  $$\lim_{y\to x} \xi_U(y)=\xi_U(x).$$ 
\end{lemma}

\proof
If $x\in |\Del(U)|$ then the result holds due to the continuity of the barycentric coordinates transformation. If $x\not\in|\Del(U)|$, since the origin $o\in |\Del(U)|$, then $||x|| \neq 0$. Therefore,  for $y$ close to $x$, $||y|| \neq 0$ and $w^y=R\frac{y}{||y||}\in\mathbb{R}^n$. 
Besides, 
$$\lim_{y\to x} w^y=w^x$$
 and therefore 
 $$\lim_{y\to x} \xi_U(y)=\xi_U(x)\ ,$$
 concluding the proof.
\qed

\begin{lemma}[Consistence]
Let $\varphi^{\scriptscriptstyle (0)}$ be the map defined in Subsection~\ref{subsubsec:maps}. 
If  $U=V$ 
and $\varphi_{\scriptscriptstyle U}^{\scriptscriptstyle (0)}(u)=\varphi^{\scriptscriptstyle (0)}(u)$ for all $u\in U$  then
\[\argmax{\varphi_{\scriptscriptstyle U}(x)}=\argmax{\varphi(x)};\mbox{ for all }x\in \Del(V)\,.\]
\end{lemma}

\proof
Let us observe that if $U=V$
and $\varphi_{\scriptscriptstyle U}^{\scriptscriptstyle (0)}(u)=\varphi^{\scriptscriptstyle (0)}(u)$ for all $u\in U$ then, for any $x\in \Del(V)$, we have that:
$$\mbox{$\varphi(x)=
\sum_{\scriptscriptstyle t\in [\![1,m]\!]
} \xi_t(x)\varphi^{\scriptscriptstyle (0)}(u^t)
=\sum_{\scriptscriptstyle t\in [\![1,m]\!]} \xi_t(x)\varphi_{\scriptscriptstyle U}^{\scriptscriptstyle (0)}(u^t)$.}$$
Then, $\varphi_{\scriptscriptstyle U}(x)=\softmax\big(\varphi(x)\big)$ and 
$\argmax{\varphi_{\scriptscriptstyle U}(x)}=\argmax{\varphi(x)}.$
\qed

One of the keys to our study is the identification of the points of ${\mathbb R}^n$ allocated inside a given simplex, with the set of all probability distributions with $n+1$ support values. 
In this way, the barycentric coordinates of a point can be seen as a probability distribution.
From this point of view, given $x\in B^n$, then $\varphi(x)$ and $\varphi_{\scriptscriptstyle U}(x)$ are both in the set $|L|$ of probability distributions with $k$ support points. 
This is why the categorical cross-entropy loss function $\Ll$ will be used to compare the similarity between $\varphi$ and $\varphi_{\scriptscriptstyle U}$.
Specifically, for $v\in V$, $\Ll$ is  defined as:
\[\mbox{$\Ll(\varphi_{\scriptscriptstyle U},\varphi,v)=-\sum_{\scriptscriptstyle h\in [\![1,k]\!]}y_h\log(s_h)$}\,,\]
where $\varphi^{\scriptscriptstyle (0)}(v)=(y_1,\dots,y_k)$ and 
$\varphi_{\scriptscriptstyle U}(v)=(s_1,\dots,s_k)$.

The following lemma establishes a specific set $U\subset V$ and a function $\hat{\varphi}_{\scriptscriptstyle U}$ such that  $\Ll(\varphi_{\scriptscriptstyle U},\varphi,v)=0$ for all $v\in V$. 

\begin{lemma}[$\Ll$-optimum simplicial map]
\label{lemma:smnn-optimo}
Let $U$ be a subset of $V$ satisfying, for all  $u\in U$, that:
\begin{enumerate}
    \item ${\varphi}_{\scriptscriptstyle U}^{\scriptscriptstyle (0)}(u)=
\varphi^{\scriptscriptstyle (0)}(u)
$, and
\item if $v\in V$ such that $\varphi^{(0)}(v)\neq \varphi^{(0)}(u)$ and $\langle v,u\rangle\in \Del(V)$ then $v\in U$.
\end{enumerate}
Then, 
\[\argmax{\varphi_{\scriptscriptstyle U}(x)}=\argmax{\varphi(x)};\mbox{ for all }x\in \Del(U)\,.\]
\end{lemma}

\proof
As proved in \cite{smnn-optimo}, 
under the assumptions stated in this lemma, we have that, for all $x\in <|\Del(U)|$:
$$\mbox{$\varphi(x)
=\sum_{\scriptscriptstyle t\in [\![1,m]\!]} \xi_t(x)\varphi_{\scriptscriptstyle U}^{\scriptscriptstyle (0)}(u^t)
$}$$
and then $\argmax{\varphi_{\scriptscriptstyle U}(x)}=\argmax{\varphi(x)}$.
\qed

Unfortunately, to compute the subset $U$ satisfying the conditions stated in 
Lemma~\ref{lemma:smnn-optimo}, we need to compute the entire triangulation $\Del(V)$ which is computationally expensive, as we have already mentioned above.

\section{Training  SMNNs}\label{sec:train}

The novel idea of this paper is to learn the function ${\varphi}_{\scriptscriptstyle U}^{\scriptscriptstyle (0)}$ for any given $U\subset V$, using the gradient descent algorithm, in order to minimize the loss function 
$\Ll(\varphi_{\scriptscriptstyle U},\varphi,v)$ for any $v\in V$.
The following result provides an expression of the gradient of  $\Ll$ in terms of the functions $\varphi_{\scriptscriptstyle U}$ and $\varphi$,
and the set $V$.

\begin{theorem}\label{th:1}
Let $U=\{u^1,\dots,u^m\}$ be a subset with $m$ elements taken from a finite set of points $V\in\R^n$ tagged with labels taken from a set of $k$ labels. Let $\varphi_{\scriptscriptstyle U}:B^n\to |L|$ and $\varphi^{(0)}:V\to W^k$.
Let us consider that
 \[\big\{\,\varphi_{\scriptscriptstyle U}^{\scriptscriptstyle (0)}(u^{t})=(p_1^t,\dots,p_k^t):\; t\in[\![1,m]\!]\,\big\}\]
is a set of variables.
Then, 
for $v\in V$,
 $$\mbox{$\frac{\partial \Ll(\varphi_{\scriptscriptstyle U},\varphi,v)}{\partial p_j^t}= (s_j-y_j)\xi_t(v)$}$$
 where $j\in[\![1,k]\!]$,
 $t\in[\![1,m]\!]$,    $\varphi^{(0)}(v)=(y_1,\dots,y_k)$
 and $\varphi_{{\scriptscriptstyle U}}(v)=(s_1,\dots,s_k)$.
\end{theorem}

\proof We have:
\begin{eqnarray*}
\mbox{$\frac{\partial \Ll(\varphi_{\scriptscriptstyle U},\varphi,v)}{\partial p_j^t}$}&=&
\mbox{$-\frac{\partial \big(\sum_{h\in [\![1,k]\!]} y_h\; \log (s_h)\big)}{\partial p_j^t}$}
\\
&=&\mbox{$-\sum_{\scriptscriptstyle h\in [\![1,k]\!]} y_h\; \frac{\partial\log (s_h)}{\partial p_j^i}$}
\\
&=&\mbox{$-\sum_{\scriptscriptstyle h\in [\![1,k]\!]} y_h\; \frac{\partial\log (s_h)}{\partial z_j}
\frac{\partial z_j}{\partial p_j^i}$}\,.
\end{eqnarray*}
Since
 $s_h=\frac{e^{z_h}}{\sum_{t\in [\![1,k]\!]} e^{z_t}}$ then 
\begin{eqnarray*}
\mbox{$\frac{\partial\log (s_h)}{\partial z_j}$}&=&
\mbox{$\frac{\partial\log \big(\frac{e^{z_h}}{\sum_{t\in [\![1,k]\!]} e^{z_t}}\big)}{\partial z_j}$}
\\
&=&\mbox{$\frac{\partial\log (e^{z_h})}{\partial z_j}-\frac{\partial\log \big(\sum_{t\in [\![1,k]\!]} e^{z_t}\big)}{\partial z_j}$}\\
&=&\mbox{$\frac{\partial z_h}{\partial z_j}-\frac{1}{\sum_{t\in [\![1,k]\!]} e^{z_t}}
\; \sum_{\scriptscriptstyle t\in [\![1,k]\!]} \frac{\partial e^{z_t}}{\partial z_j}$}
\\
&=&\mbox{$\delta_{hj}-\frac{ e^{z_j}}{\sum_{t\in [\![1,k]\!]} e^{z_t}}$}
=\delta_{hj}-s_j\,.
\end{eqnarray*}
Besides, since
 $z_j=\sum_{\scriptscriptstyle h\in [\![1,m]\!]}\xi_h(v) p_j^h$ then
$$\mbox{$
\frac{\partial z_r}{\partial p_j^t}=
\sum_{h\in [\![1,m]\!]} \xi_h(v)\frac{\partial p_r^h}{\partial p_j^t}=\xi_t(v)$}\,.$$
Finally, 
\begin{eqnarray*}
\mbox{$\frac{\partial\Ll(\psi,\varphi,v)}{\partial p_j^t}$}&=&
\mbox{$-\sum_{\scriptscriptstyle  h\in [\![1,k]\!]} y_h(\delta_{hj}-s_j)\xi_t(v)$}
\\
&=&
\mbox{$-\xi_t(v)\big(\sum_{\scriptscriptstyle h\in [\![1,k]\!]} y_h\delta_{hj}-s_j\sum_{\scriptscriptstyle h\in [\![1,k]\!]}y_h\big)$}
\\
&=&
(s_j-y_j)\xi_t(v)\,.
\end{eqnarray*}
\qed

Let us now see how we add trainability to the SMNN $\N_{\varphi_{\scriptscriptstyle U}}$ induced by $\varphi_{\scriptscriptstyle U}$. Let $V$ be the training set and let $U$ be a support set lying in the same space as $V$.
First, assuming that $U=\{u^1,\dots,u^m\}$ has $m$ elements, then $\N_{\varphi_{\scriptscriptstyle U}}$ is a multiclass perceptron 
with an input layer with $m$ neurons that predicts the $h$-th label for $h\in [\![1,k]\!]$ using the formula:
\[\mbox{$\N_{\varphi_{\scriptscriptstyle U}}(x)=
\softmax\big(\tilde{\M}\cdot \xi_U(x)\big)
$}\]
where $\tilde{\M}=(p_j^t)_{j\in [\![1,k]\!],\, t\in[\![1.m]\!]} $ is a matrix of weights and $\xi_U(x)\in \R^{m}$ is obtained from the barycentric coordinates of $x\in B^n$ as in Section~\ref{sec:unknown}. 
Let us observe that 
\[\softmax\big(\tilde{\M}\cdot \xi_U(x)\big) \in |L|.\]
The idea is to modify the initial values of \[\mbox{$\varphi^{\scriptscriptstyle(0)}_{\scriptscriptstyle U}(u^t)=(p_1^t,\dots,p_k^t)\;$
for $u^t\in U$ and $t\in[\![1,m]\!]$,}\]
in order to obtain new values for $\N_{\varphi_{\scriptscriptstyle U}}
(v)$ for $v\in V$ in a way that the error $\Ll(\N_{\varphi_{\scriptscriptstyle U}},\varphi,v)$ decreases. 
We will do it by avoiding recomputing $\Del (U)$ or the barycentric coordinates $(b_0(v),\dots, b_n(v))$ for each $v\in V$ during the training process.

In this way, given $v\in V$, if $v\in|\Del(U)|$, we compute the maximal simplex $\sigma=\langle u^{i_0},\dots,u^{i_n}\rangle\in\Del(U)$ such that $v\in|\sigma|$  and   $i_h\in [\![1,m]\!] $ for $h\in [\![0,n]\!]$. If $v\not\in|\Del(U)|$,   we compute $w\in S^n$ and the simplex $\sigma=\langle w, u^{i_1},\dots,u^{i_n} \rangle$ 
such that $v\in|\sigma|$  and   $i_h\in [\![1,m]\!] $ for $h\in [\![1,n]\!]$.
Then we compute the barycentric coordinates $b(v)$ of $v$ with respect to $\sigma$ and the point $\xi_U(x)=(\xi_1(x),\dots,\xi_m(x))\in \R^m$ as in Section~\ref{sec:unknown}.

Using the gradient descent algorithm, we update the variables $p_j^{t}$ for $j\in [\![1,k]\!]$ and $t\in [\![1,m]\!]$ as follows:
\[\mbox{$p_j^{t}:=p_j^{t}-\eta\frac{\partial\Ll(\N_{\varphi_U,}
\varphi,v)}{\partial p_j^{t}}
=
p_j^{t}-\eta(s_j-y_j)\xi_{t}(v)$.}\]

\begin{figure}[]
    \centering
    \includegraphics[width=0.8\linewidth]{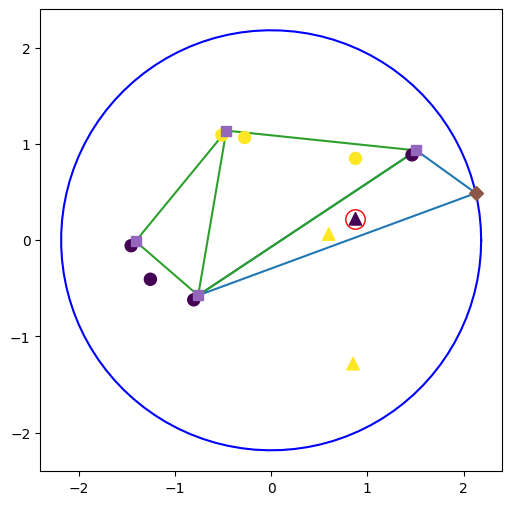}
    \caption{Two-dimensional 
    synthetic dataset with points divided into two classes: blue and yellow. Triangle-shaped points belong to the test set and square-shaped points belong to the support set $U$.
    The diamond-shaped point is the vertex on the hypersphere (the blue circumference) used to classify the triangle-shaped 
    point $v$ (surrounded by a small red circumference) outside the triangulation. }
    \label{fig:example_methodology}
\end{figure}

An illustrative picture of the role of each point in a simple two-dimensional binary classification problem is provided in Figure \ref{fig:example_methodology}.
The pseudocode of the method to train SMNNs using Stochastic Gradient Descent
 is provided in Algorithm~\ref{algo:SMNN_train}.

\begin{algorithm}
\caption{Pseudocode of the proposed method to train SMNNs using SGD.}
\label{algo:SMNN_train}
\begin{algorithmic}
\State {\bf Input:} 
Dataset $V\subset \mathbb{R}^n$ surrounded by a hypersphere of radius $R$ and a model $\mathcal{N}_{\varphi_U}$.
\State {\bf Parameter:} $\eta>0$
\State {\bf Output:} The trained model 
$\N_{\varphi_U}$
\State {\bf init} $\tilde{\M}$, the matrix of weights of $\N_{\varphi_U}$
\For{$v\in V$}
\State {\bf compute} $\xi_U(v)$ as in Section~\ref{sec:unknown}
\For{each column $p$ of $\M$}
\State  $p:=p-\eta\frac{\partial\Ll(
\N_{\varphi_U},
\varphi,v)}{\partial p}$
\EndFor
\EndFor
\State $\N_{\varphi_U}(x):=\softmax (\tilde{\M}\cdot \xi_U(x))$
\end{algorithmic}
\end{algorithm}

\begin{figure*}[]
    \centering
    \includegraphics[width=\linewidth]{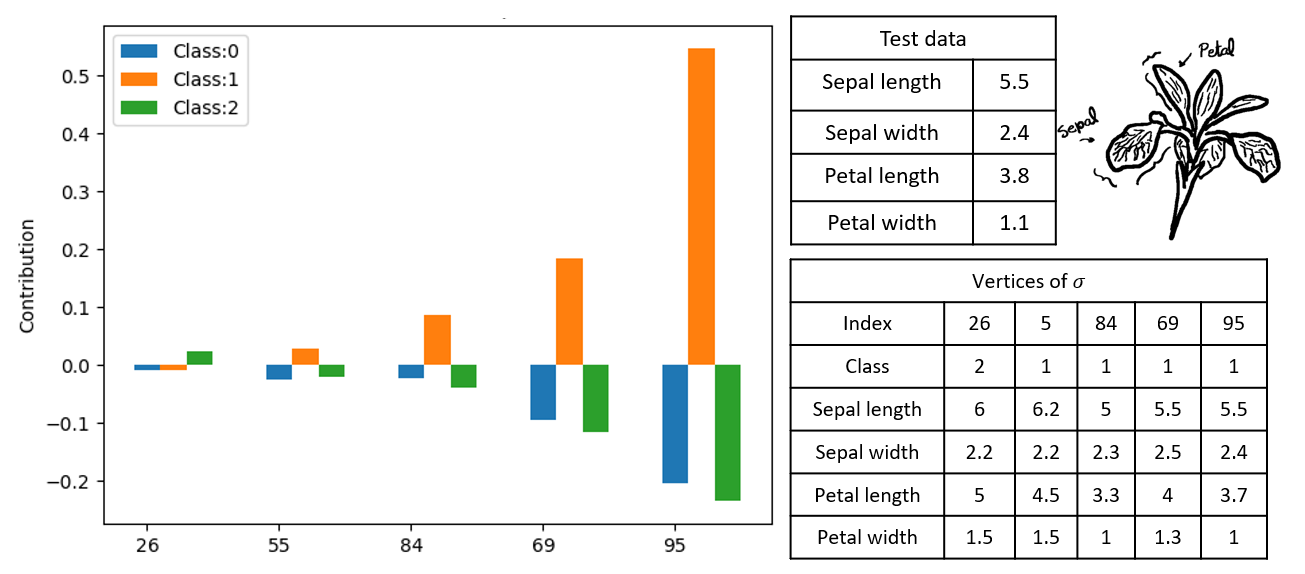}
    \caption{Table with the five flowers taken from the training set 
    that influence the classification of a given flower in the test set, i.e., the vertices of $\sigma$. The SMNN was trained for $1000$ epochs and $R\approx 6.16$.}
    \label{fig:example_explain_iris}
\end{figure*}

\section{Explainability}\label{sec:Explain}
In this section, we provide insight into the explainability capability of SMNNs. In the literature, many different approaches can be found to {\it what is an explanation} of the prediction of an AI model. In our case, explainability will be provided based on similarities and dissimilarities of the point $x$ to be explained with the points corresponding to the vertices of the simplex $\sigma$ containing it. Based on this idea, the barycentric coordinates of $x$
with respect to $\sigma$
can be considered indicators of how much a vertex of $\sigma$
will contribute to the prediction of $x$.
Specifically,
the barycentric coordinates of $x$ multiplied by the evaluation of
the trained map $\varphi_{\scriptscriptstyle U}^{\scriptscriptstyle (0)}$  
at the vertices of $\sigma$ encodes the contribution of each vertex of $\sigma$ 
to the label assigned to $x$ by the SMNN.

As an illustration, consider the Iris dataset\footnote{https://archive.ics.uci.edu/ml/datasets/iris} as a toy example and split it into a training set ($75\%$) and a test set ($25\%$). Since we focus on this section on explainability, let us take $U=V$, containing 112 points.
Then,  initialize   $p_j^t$  with a random value in $[0,1]$,  for
$j\in[\![1,4]\!]$ and $t\in[\![1,112]\!]$.

After the training process,
the SMNN reached $92\%$ accuracy and $0.5$ loss value on the test set.
Once the SMNN is trained,
we may be interested in determining why a certain output is given for a specific point $x$ in the test set. 

As mentioned above, the explanation for why the SMNN assigns a label to $x$ is based on the labels of the vertices of the simplex of $\Del(U)$ containing $x$.
Therefore, the first step is to find the maximal simplex $\sigma$
that contains the point $x$
to be explained. 
As an example, in Figure~\ref{fig:example_explain_iris}, the point 
$x=(5.5,2.4,3.8,1.1)\in\R^4$ in the test set is chosen to be explained, 
to which the SMNN predicts to assign class 2.
The coordinates of the five
vertices ($u^{26}$, $u^{55}$, $u^{69}$, $u^{84}$ and $u^{95}$) of the simplex $\sigma$ containing $x$
together with the classes they have been assigned
are shown in the table at the bottom of Figure~\ref{fig:example_explain_iris}.
The contribution of the class assigned to each vertex of $\sigma$ to 
the class assigned to $x$ by the SMNN is displayed in the bar chart, and is measured in terms of $p_j^t\cdot \xi_t(x) $ for $j\in [\![0,2]\!]$ and $t\in\{26,55,69,84,95\}$.
Let us notice that the contributions can be positive or negative. 
For example, 
the vertex with index 95 with the greatest influence 
on the prediction 
negatively affected the classification of $x$ corresponding to the first and third class, but positively to the second class,
which is the correct classification. Let us note that the Euclidean distance between points is not the only factor making a vertex of $\sigma$ contribution greater. That is,
even if two vertices
are equally close to the point to be explained, they do
not contribute the same. 
For
example, the 
vertices $84$ and $95$ are 
similarly close to the test point, but their contribution is very different in magnitude.

\section{Discussion and Limitations}\label{sec:disc}

Let us remark that SMNNs are in between an instance-based method and a multilayer perceptron. The previous definition of SMNNs (\cite{DBLP:journals/nn/Paluzo-HidalgoG20,smnn}) shared advantages with instance-based methods such as the k-Nearest Neighbour Algorithm. Some of the advantages are: there was no training phase, it handles easily complex decision boundaries by barycentric subdivisions, it is effective in low-dimensional spaces, it adapts well to local patterns in the data, and the decision-making is transparent and interpretable.  However, it was computationally expensive for large high-dimensional datasets due to the Delaunay triangulation computation and the required convex polytope,
it suffered from overfitting and lacked generalization ability. The proposed update in this paper provides a substitute for the convex polytope which reduces the number of points needed to compute the Delaunay triangulation.
Delaunay computation is also less expensive thanks to the use of a support set $U$. 

Nevertheless, one of the main limitations of SMNNs is the need for an input triangulable space. Hence, structured data such as images need to be embedded by, for example, applying a dimensionality reduction technique such as UMAP \cite{umap} so that the dataset is a point cloud. 

Regarding the explainability of the model, it is not a feature-based explainability such as \cite{10.1007/978-3-031-42941-5_28, Nguyen2019} and, hence, it does not provide direct insight on the importance of the features of the input data. However, it provides instance-based explainability \cite{kong2021understanding}. When predicting input data, the vertices of the simplex where it belongs and their contribution to the classification give a similarity measure inferred by the SMNN which is understandable by an
expert.

\begin{table}[]
\begin{center}
\caption{Accuracy score and loss values obtained after training both the SMNN and a feed-forward neural network (FFNN). 
The experiments were repeated $100$ times and the results provided are the mean of the accuracy values of the repetitions. 
The size $m$ of the subset considered to compute the Delaunay triangulation also varies in each experiment depending on a parameter $\kappa$.
 The FFNN  is composed of two hidden layers of size $32$ and $16$, respectively, with ReLu activation functions and an output layer with a softmax activation function. The datasets used are synthetic binary-class datasets with $n=2,3,4,5$ features.  }
\begin{tabular}{c|cccc|cc|}
\cline{2-7}                                         & \multicolumn{4}{c|}{SMNN}  & \multicolumn{2}{c|}{FFNN}    \\ \hline
\multicolumn{1}{|c|}{$n$}      & \multicolumn{1}{c|}{$\kappa$} & \multicolumn{1}{c|}{$m$} & \multicolumn{1}{c|}{Acc.}                       & Loss                      & \multicolumn{1}{c|}{Acc.}                                        & Loss                                  \\ \hline
\multicolumn{1}{|c|}{\multirow{4}{*}{2}} & \multicolumn{1}{c|}{1000}          & \multicolumn{1}{c|}{3560}        & \multicolumn{1}{c|}{0.87}      & 0.64      & \multicolumn{1}{c|}{\multirow{4}{*}{0.91}}      & \multirow{4}{*}{0.23} \\ \cline{2-5}
\multicolumn{1}{|c|}{}                   & \multicolumn{1}{c|}{100}           & \multicolumn{1}{c|}{1282}        & \multicolumn{1}{c|}{0.90}      & 0.51      & \multicolumn{1}{c|}{}                                                   &                                               \\ \cline{2-5}
\multicolumn{1}{|c|}{}                   & \multicolumn{1}{c|}{50}            & \multicolumn{1}{c|}{626}         & \multicolumn{1}{c|}{0.9 }   & 0.42  & \multicolumn{1}{c|}{}                                                   &                                               \\ \cline{2-5}
\multicolumn{1}{|c|}{}                   & \multicolumn{1}{c|}{10}            & \multicolumn{1}{c|}{53}          & \multicolumn{1}{c|}{0.87 }  & 0.33   & \multicolumn{1}{c|}{}                                                   &                                               \\ \hline
\multicolumn{1}{|c|}{\multirow{4}{*}{3}} & \multicolumn{1}{c|}{1000}          & \multicolumn{1}{c|}{3750}        & \multicolumn{1}{c|}{0.76 } & 0.66  & \multicolumn{1}{c|}{\multirow{4}{*}{0.8}} & \multirow{4}{*}{0.61}               \\ \cline{2-5}
\multicolumn{1}{|c|}{}                   & \multicolumn{1}{c|}{100}           & \multicolumn{1}{c|}{3664}        & \multicolumn{1}{c|}{0.76}      & 0.66   & \multicolumn{1}{c|}{}                                                   &                                               \\ \cline{2-5}
\multicolumn{1}{|c|}{}                   & \multicolumn{1}{c|}{50}            & \multicolumn{1}{c|}{3252}        & \multicolumn{1}{c|}{0.77 } & 0.65   & \multicolumn{1}{c|}{}                                                   &                                               \\ \cline{2-5}
\multicolumn{1}{|c|}{}                   & \multicolumn{1}{c|}{10}            & \multicolumn{1}{c|}{413}         & \multicolumn{1}{c|}{0.81 } & 0.5      & \multicolumn{1}{c|}{}                                                   &                                               \\ \hline
\multicolumn{1}{|c|}{\multirow{4}{*}{4}} & \multicolumn{1}{c|}{50}            & \multicolumn{1}{c|}{3728}        & \multicolumn{1}{c|}{0.69 }  & 0.67  & \multicolumn{1}{c|}{\multirow{4}{*}{0.72 }} & \multirow{4}{*}{0.69}               \\ \cline{2-5}
\multicolumn{1}{|c|}{}                   & \multicolumn{1}{c|}{10}            & \multicolumn{1}{c|}{1410}        & \multicolumn{1}{c|}{0.73 } & 0.64  & \multicolumn{1}{c|}{}                                                   &                                               \\ \cline{2-5}
\multicolumn{1}{|c|}{}                   & \multicolumn{1}{c|}{5}             & \multicolumn{1}{c|}{316}         & \multicolumn{1}{c|}{0.73 } & 0.57  & \multicolumn{1}{c|}{}                                                   &                                               \\ \cline{2-5}
\multicolumn{1}{|c|}{}                   & \multicolumn{1}{c|}{2}             & \multicolumn{1}{c|}{26}          & \multicolumn{1}{c|}{0.72 } & 0.56  & \multicolumn{1}{c|}{}                                                   &                                               \\ \hline
\multicolumn{1}{|c|}{\multirow{4}{*}{5}} & \multicolumn{1}{c|}{50}            & \multicolumn{1}{c|}{3743}        & \multicolumn{1}{c|}{0.77 } & 0.66  & \multicolumn{1}{c|}{\multirow{4}{*}{0.8 }}  & \multirow{4}{*}{0.91}               \\ \cline{2-5}
\multicolumn{1}{|c|}{}                   & \multicolumn{1}{c|}{10}            & \multicolumn{1}{c|}{1699}        & \multicolumn{1}{c|}{0.81} & 0.63  & \multicolumn{1}{c|}{}                                                   &                                               \\ \cline{2-5}
\multicolumn{1}{|c|}{}                   & \multicolumn{1}{c|}{5}             & \multicolumn{1}{c|}{323}         & \multicolumn{1}{c|}{0.8 }  & 0.52  & \multicolumn{1}{c|}{}                                                   &                                               \\ \cline{2-5}
\multicolumn{1}{|c|}{}                   & \multicolumn{1}{c|}{2}             & \multicolumn{1}{c|}{17}            & \multicolumn{1}{c|}{0.74 }                                  &       0.53                             & \multicolumn{1}{c|}{}                                                   &                                               \\ \hline
\end{tabular}
\label{table:synthetic}
\end{center}
\end{table}

\section{Experiments}\label{sec:experiments}

In this section, we provide experiments that show the performance of SMNNs.
In all the experiments,
we split the given dataset into a training set and a test set composed of 75\% and 25\% of the dataset, respectively. 
The datasets used for experimentation are (1) a two-dimensional binary classification spiral synthetic dataset, and (2) dimension-varying binary classification synthetic datasets composed of different noisy clusters for each class (we refer to \cite{Guyon2003DesignOE} for a specific description of how data is generated). All experiments were developed using a 3.70 GHz AMD Ryzen 9 5900X 12-Core Processor.

In the first two experiments,
$\varepsilon$-representative subsets of the training set are used as the support set $U$ for different values of $\varepsilon$. 
The notion of $\varepsilon$-representative sets was introduced in \cite{representative}. 
Specifically, a support set $U$ is $\varepsilon$-representative  of a set $V$ if, for any $v\in V$, there exists $u\in U$ such that the distance between $u$ and $v$ is less than $\varepsilon$.

Let us now describe the methodology of each experiment.

\begin{figure}
     \centering
     \begin{subfigure}[]{0.33\textwidth}
         \centering
         \includegraphics[width=\textwidth]{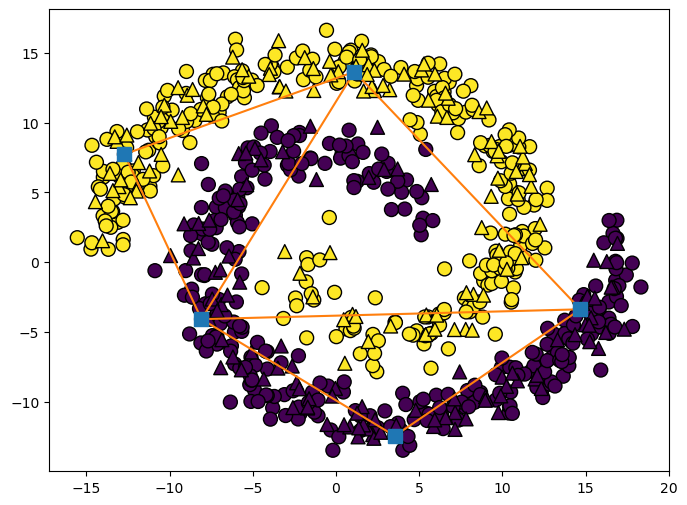}
         \caption{Using a support set of $5$ 
         points, the SMNN reaches $80\%$  test accuracy.}
         \label{fig:spiral1}
     \end{subfigure}
     \begin{subfigure}[]{0.33\textwidth}
         \centering
         \includegraphics[width=\textwidth]{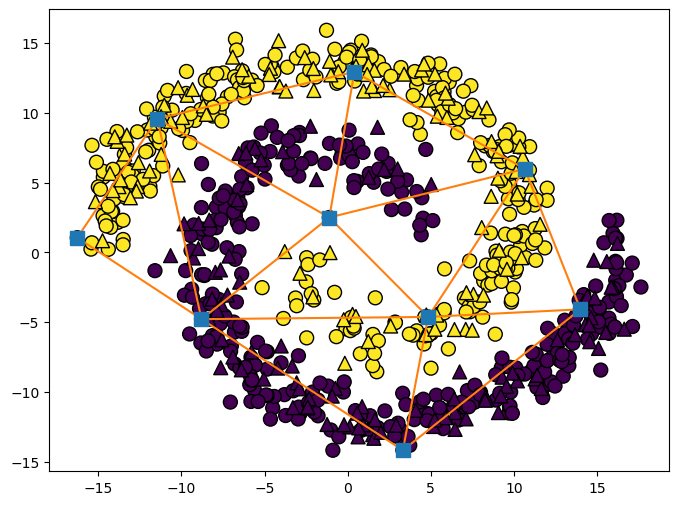}
         \caption{Using a support set of $9$
         points, the SMNN reaches $93\%$ test accuracy.}
         \label{fig:spiral2}
     \end{subfigure}
     \begin{subfigure}[]{0.33\textwidth}
         \centering
         \includegraphics[width=\textwidth]{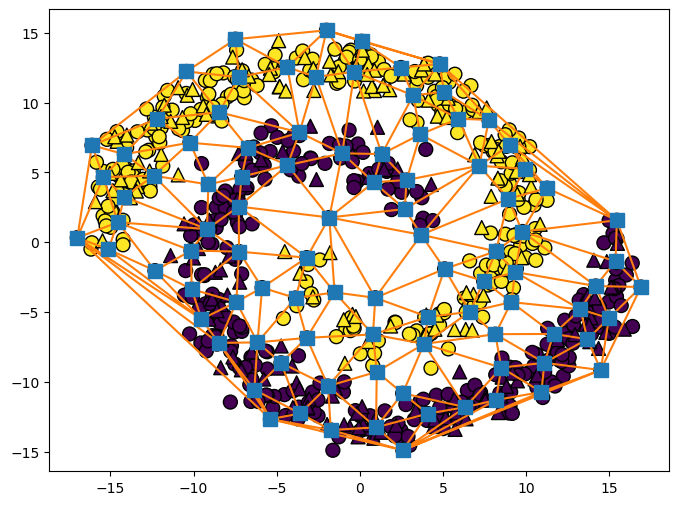}
         \caption{Using a support set of $95$ 
         points, the SMNN reaches $99\%$  test accuracy.}
         \label{fig:spiral3}
     \end{subfigure}
        \caption{Spiral dataset for binary classification. On each figure, the points in the support set are square-shaped, the points in the test set are triangle-shaped and the points in the training set are circle-shaped. }
        \label{fig:spiral_dataset}
\end{figure}

{\bf First experiment:} we consider a two-dimensional spiral dataset for binary classification
composed of $400$ two-dimensional points. 
We selected three different values of $\varepsilon$ 
obtaining three $\varepsilon$-representative sets (the support sets) of size 5, 9 and 95, respectively. In Figure~\ref{fig:spiral_dataset}, the spiral dataset and the three different support sets with the associated Delaunay triangulation are shown. 
In this case, we observed that the accuracy of the SMNNs increases with the size of the support set. We can also appreciate that the topology of the dataset is characterized by the support set, which we guess is responsible for the successful classification.

\begin{figure}
\centering
\begin{subfigure}[]{0.3\textwidth}  \centering\includegraphics[width=\textwidth]{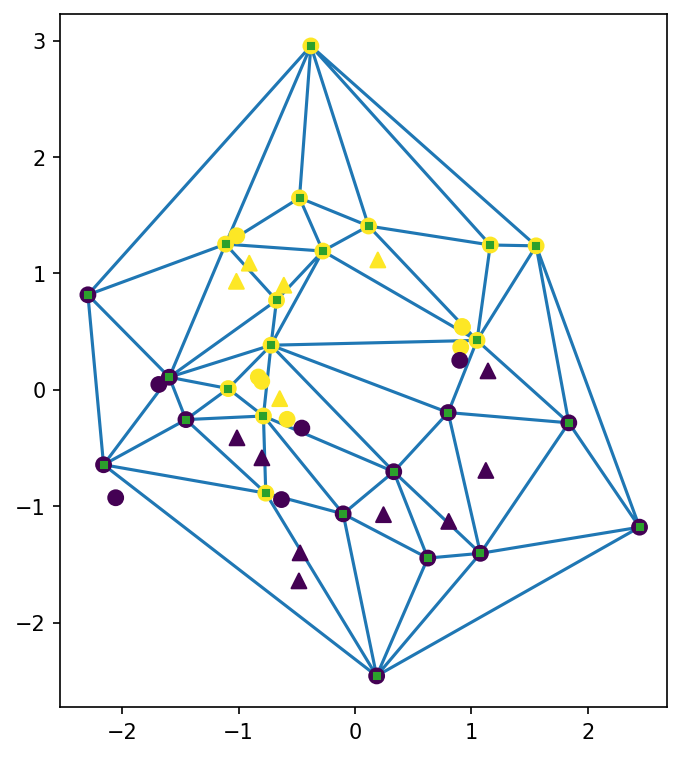}
    \caption{$\varepsilon$-Representative set.}
\end{subfigure}
\begin{subfigure}[]{0.3\textwidth}  \centering\includegraphics[width=\textwidth]{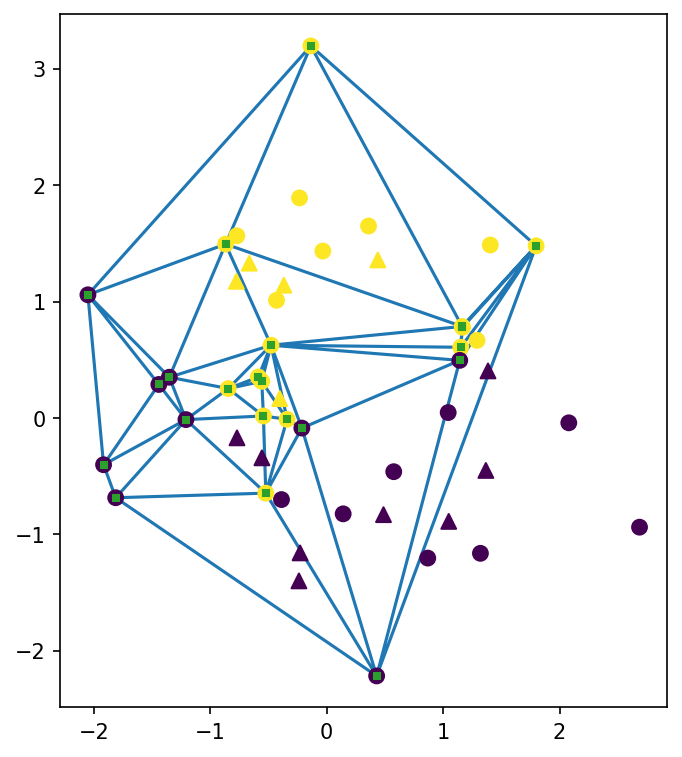}
    \caption{Outlier-robust representative landmark set.}
\end{subfigure}
\begin{subfigure}[]{0.3\textwidth}
\centering
    \includegraphics[width=\textwidth]{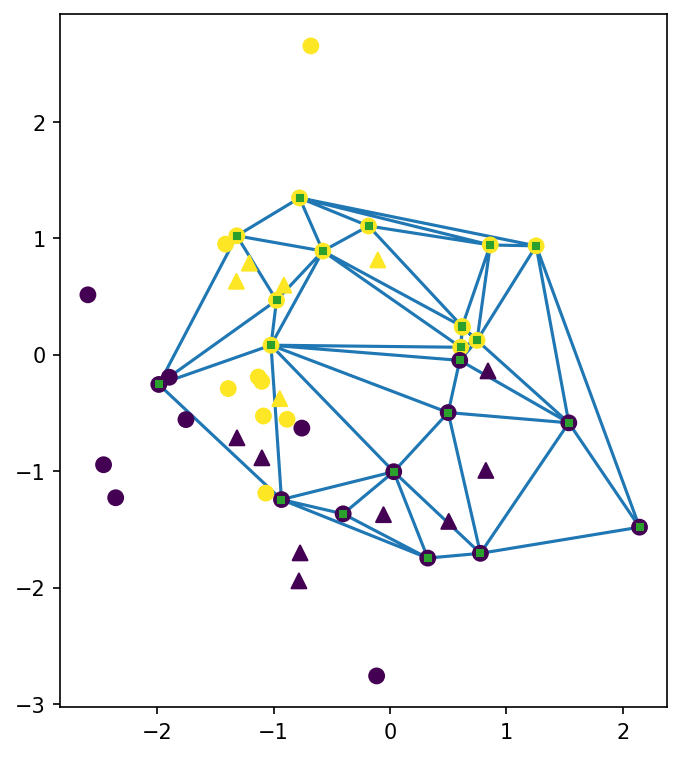}
    \caption{Outlier-robust vital landmark set.}
\end{subfigure}
\caption{Examples of support sets obtained using the different methods proposed in the third experiment.}
\label{fig:supportsets}
\end{figure}

{\bf Second experiment:} we consider four synthetic datasets of size $5000$
using the adaptation of \cite{Subramanian2003DesignOE} in 
the scikit-learn \cite{scikit-learn} implementation.
 The four datasets generated have, respectively, $n=2,3,4$ and $5$ features.
 Then, the corresponding training set $V$ obtained from each dataset 
 and a fully connected two-hidden-layer feed-forward ($32\times 16$) neural network (FNNN) with ReLU activation functions
 were considered.

To train the SMNN, we use
four different $\varepsilon$-representative sets of each $V$. 
In our experiments, the different values of $\varepsilon$ are calculated as the maximum distance from the origin to the 
farthest
point in the dataset plus $\frac{1}{2}$ and divided by a parameter $\kappa$. The different values for $\kappa$ considered are $1000,100,50,10$.
Using the different values of $\kappa$ we then obtain support sets $U$ of different sizes $m$. For example, for $n=2$ and $k=1000$, we obtain a support set $U$ of size $m=3560$.
The sizes $m$ of the support sets $U$ generated for $n=2,3,4,5$ and $\kappa=1000,100,50,10$ are provided in Table~\ref{table:synthetic}. 

First, the SMNN was trained using the gradient descent algorithm and the cross-entropy loss function during $500$ epochs for $n=2,3,4,5$ and $\kappa=1000,100,50,10$. 
Besides, the two-hidden-layer feed-forward neural network FFNN was trained using the Adam training algorithm \cite{adam} for $n=2,3,4,5$.
Both training processes were carried out for $500$ epochs.
The accuracy and loss results provided in Table~\ref{table:synthetic} are the mean of $100$ repetitions.  
We can observe that both the SMNN and the FFNN have similar performance, but the SMNN generally reaches lower loss values. The variance in the results was on the order of $10^{-8}$ to $10^{-5}$ in the case of the SMNN and of $10^{-5}$ to $10^{-2}$ in the case of the FFNN.

{\bf Third experiment:} we compare the performance of the SMNN depending on the choice of the support set $U$.
To do this, we applied two different methods to choose $U$. 
On the one hand, we use the $\varepsilon$-representative sets previously computed in the first experiment for different values of $\kappa$. On the other hand, we use the two outlier-robust subsampling methods presented in~\cite{JMLR:v24:21-1526} for different values of Topological Density (TD). 
In Figure~\ref{fig:supportsets}, examples of different support sets computed using the three different approaches are shown. Let us remark that the outlier-robust subsampling method can be tuned to keep outliers, which the authors call \textit{vital landmark} set, or not, obtaining a \textit{representative landmark} set. The two methods were tested on synthetic datasets composed of $1000$ points. The SMNN was trained for $1000$ epochs using the $75\%$ of the dataset and tested in the remaining $25\%$.
The number of features $n$ of each dataset considered, the size $m$ of each support set computed, and the mean accuracy and loss results of $5$ repetitions when training the SMNN are provided in Table~\ref{table:outlier}. 
In Table~\ref{table:time}, the time for the computation of the barycentric coordinates is provided. As we can see, the time of execution does not have to be directly related to the size of the support set and it may increase if many of the points to be evaluated are outside the Delaunay triangulation of $U$.
In this experiment, the results suggest that $\varepsilon$-representative datasets provide better results than the other support sets. 
However, thorough theoretical studies should be developed to confirm the last statement. Such a study is outside of the scope of this paper.

\begin{table*}[]
\begin{center}
\caption{Accuracy score and loss values obtained after training the SMNN. 
The size $m$ of the subset considered to compute the Delaunay triangulation varies in each experiment depending on a parameter $\kappa$ (for $\varepsilon$-representative sets) or a parameter TD (for landmark sets).
 The datasets used are synthetic binary-class datasets 
with $n=2,3,4$ features.  }
\begin{tabular}{|cc|c|ccc|ccc|ccc|c|}
\hline
\multicolumn{2}{|c|}{\multirow{2}{*}{\bf Sampling method}}                                     & \multirow{2}{*}{\bf Parameter} & \multicolumn{3}{c|}{\bf $n=2$}                                    & \multicolumn{3}{c|}{\bf $n=3$}                                    & \multicolumn{3}{c|}{\bf $n=4$}                                    \\ \cline{4-12} 
\multicolumn{2}{|c|}{}                                                                     &                            & \multicolumn{1}{c|}{\bf $m$}   & \multicolumn{1}{c|}{\bf Acc.} & {\bf Loss} & \multicolumn{1}{c|}{\bf $m$}   & \multicolumn{1}{c|}{\bf Acc.} & {\bf Loss} & \multicolumn{1}{c|}{\bf $m$}   & \multicolumn{1}{c|}{\bf Acc.} & {\bf Loss} \\ \hline
\multicolumn{2}{|c|}{\multirow{3}{*}{Representative sets}}                                 & $\kappa=10$                       & \multicolumn{1}{c|}{48}  & \multicolumn{1}{c|}{0.94} & 0.22 & \multicolumn{1}{c|}{109} & \multicolumn{1}{c|}{0.94} & 0.29 & \multicolumn{1}{c|}{407} & \multicolumn{1}{c|}{0.9}  & 0.53 \\ \cline{3-12} 
\multicolumn{2}{|c|}{}                                                                     & $\kappa=50$                       & \multicolumn{1}{c|}{371} & \multicolumn{1}{c|}{0.94} & 0.47 & \multicolumn{1}{c|}{730} & \multicolumn{1}{c|}{0.95} & 0.53 & \multicolumn{1}{c|}{748} & \multicolumn{1}{c|}{0.87} & 0.6  \\ \cline{3-12} 
\multicolumn{2}{|c|}{}                                                                     & $\kappa=100$                      & \multicolumn{1}{c|}{570} & \multicolumn{1}{c|}{0.93} & 0.54 & \multicolumn{1}{c|}{747} & \multicolumn{1}{c|}{0.96} & 0.56 & \multicolumn{1}{c|}{750} & \multicolumn{1}{c|}{0.87} & 0.6  \\ \hline
\multicolumn{1}{|c|}{\multirow{8}{*}{ORS}} & \multirow{4}{*}{Representative landmark sets} & TD=0.1                     & \multicolumn{1}{c|}{75}  & \multicolumn{1}{c|}{0.79} & 0.49 & \multicolumn{1}{c|}{75}  & \multicolumn{1}{c|}{0.9}  & 0.35 & \multicolumn{1}{c|}{75}  & \multicolumn{1}{c|}{0.85} & 0.46 \\ \cline{3-12} 
\multicolumn{1}{|c|}{}                     &                                               & TD=0.4                     & \multicolumn{1}{c|}{300} & \multicolumn{1}{c|}{0.85} & 0.57 & \multicolumn{1}{c|}{300} & \multicolumn{1}{c|}{0.86} & 0.5  & \multicolumn{1}{c|}{300} & \multicolumn{1}{c|}{0.87} & 0.62 \\ \cline{3-12} 
\multicolumn{1}{|c|}{}                     &                                               & TD=0.6                     & \multicolumn{1}{c|}{450} & \multicolumn{1}{c|}{0.86} & 0.52 & \multicolumn{1}{c|}{450} & \multicolumn{1}{c|}{0.89} & 0.52 & \multicolumn{1}{c|}{450} & \multicolumn{1}{c|}{0.86} & 0.55 \\ \cline{3-12} 
\multicolumn{1}{|c|}{}                     &                                               & TD=0.8                     & \multicolumn{1}{c|}{600} & \multicolumn{1}{c|}{0.89} & 0.56 & \multicolumn{1}{c|}{600} & \multicolumn{1}{c|}{0.92} & 0.52 & \multicolumn{1}{c|}{600} & \multicolumn{1}{c|}{0.86} & 0.58 \\ \cline{2-12} 
\multicolumn{1}{|c|}{}                     & \multirow{4}{*}{Vital landmark sets}          & TD=0.1                     & \multicolumn{1}{c|}{75}  & \multicolumn{1}{c|}{0.93} & 0.27 & \multicolumn{1}{c|}{75}  & \multicolumn{1}{c|}{0.84} & 0.4  & \multicolumn{1}{c|}{75}  & \multicolumn{1}{c|}{0.88} & 0.37 \\ \cline{3-12} 
\multicolumn{1}{|c|}{}                     &                                               & TD=0.4                     & \multicolumn{1}{c|}{300} & \multicolumn{1}{c|}{0.93} & 0.37 & \multicolumn{1}{c|}{300} & \multicolumn{1}{c|}{0.88} & 0.46 & \multicolumn{1}{c|}{300} & \multicolumn{1}{c|}{0.9}  & 0.45 \\ \cline{3-12} 
\multicolumn{1}{|c|}{}                     &                                               & TD=0.6                     & \multicolumn{1}{c|}{450} & \multicolumn{1}{c|}{0.91} & 0.44 & \multicolumn{1}{c|}{450} & \multicolumn{1}{c|}{0.96} & 0.44 & \multicolumn{1}{c|}{450} & \multicolumn{1}{c|}{0.89} & 0.53 \\ \cline{3-12} 
\multicolumn{1}{|c|}{}                     &                                               & TD=0.8                     & \multicolumn{1}{c|}{600} & \multicolumn{1}{c|}{0.93} & 0.5  & \multicolumn{1}{c|}{600} & \multicolumn{1}{c|}{0.95} & 0.52 & \multicolumn{1}{c|}{600} & \multicolumn{1}{c|}{0.87} & 0.57 \\ \hline
\end{tabular}
\label{table:outlier}
\end{center}
\end{table*}

\begin{table}[]
\caption{Time in seconds for the $\xi_U(x)$ computation for the experiments in Table~\ref{table:outlier}. The values are the mean of 5 iterations. Let us remark that higher values can be expected when increasing the number of points outside the Delaunay triangulation of the support set.}
    \centering
\resizebox{\columnwidth}{!}{
    \begin{tabular}{|cc|c|c|c|c|}
\hline
\multicolumn{2}{|c|}{\textbf{Sampling method}}                                             & \textbf{Parameter} & \textbf{$n=2$} & \textbf{$n=3$} & \textbf{$n=4$} \\ \hline
\multicolumn{2}{|c|}{\multirow{3}{*}{Representative sets}}                                 & $\kappa=10$               & 0.08         & 0.14         & 0.54         \\ \cline{3-6} 
\multicolumn{2}{|c|}{}                                                                     & $\kappa=50$               & 0.12         & 0.22         & 0.79         \\ \cline{3-6} 
\multicolumn{2}{|c|}{}                                                                     & $\kappa=100$              & 0.14         & 0.24         & 0.79         \\ \hline
\multicolumn{1}{|c|}{\multirow{8}{*}{ORS}} & \multirow{4}{*}{Representative landmark sets} & TD=0.1             & 1.04         & 2.89         & 21.75        \\ \cline{3-6} 
\multicolumn{1}{|c|}{}                     &                                               & TD=0.4             & 0.33         & 0.29         & 7            \\ \cline{3-6} 
\multicolumn{1}{|c|}{}                     &                                               & TD=0.6             & 0.12         & 0.37         & 3.03         \\ \cline{3-6} 
\multicolumn{1}{|c|}{}                     &                                               & TD=0.8             & 0.1          & 0.23         & 1.29         \\ \cline{2-6} 
\multicolumn{1}{|c|}{}                     & \multirow{4}{*}{Vital landmark sets}          & TD=0.1             & 0.17         & 3.22         & 3.87         \\ \cline{3-6} 
\multicolumn{1}{|c|}{}                     &                                               & TD=0.4             & 0.15         & 2.31         & 1.69         \\ \cline{3-6} 
\multicolumn{1}{|c|}{}                     &                                               & TD=0.6             & 0.08         & 0.53         & 1.17         \\ \cline{3-6} 
\multicolumn{1}{|c|}{}                     &                                               & TD=0.8             & 0.15         & 0.27         & 0.82         \\ \hline
\end{tabular}
}
    \label{table:time}
\end{table}

\section{Conclusions}
The balance between efficiency and explainability will be one of the major problems of AI in the next years. Although AI models based on network architectures and backpropagation algorithms are currently among the most successful models, they are far from providing a human-readable explanation of their outputs. On the other hand, simpler models not based on gradient descent methods usually do not provide a comparable level of performance. In this way, a trainable version of SMNNs provides a new step in filling the gap between both approaches.

Simplicial map neural networks provide a combinatorial approach to artificial intelligence. Its simplicial-based definition provides nice properties, such as easy construction and robustness capability against adversarial examples. 

In this work, we have extended its definition to provide a trainable version of this architecture. The training process is based on a local search and links this model based on simplices with the most efficient methods in AI. Moreover, we have demonstrated in this paper that such a simplicial-based construction provides a human-understandable explanation of the decision.

The ideas presented in this paper can be extended in many different ways. In future work, we intend to study less data-dependent approaches so that the Delaunay triangulation is needless. Besides, this architecture should be extended to Deep Learning models so that it can be applied to more complex classification problems such as image classification and provides extra explainability to Deep Learning models.

\section*{Code availability}

The code is available in the GitHub repository: \url{https://github.com/Cimagroup/TrainableSMNN}.

\section{Aknowledgments}

The work was supported in part by the European Union HORIZON-CL4-2021-HUMAN-01-01 under grant agreement 101070028 (REXASI-PRO) and by  
TED2021-129438B-I00 / AEI/10.13039/501100011033 / Unión Europea NextGenerationEU/PRTR

{\small
\bibliographystyle{elsarticle-num-names}
\bibliography{egbib}
}

\end{document}